\documentclass{article} 
\usepackage{iclr2024_conference,times}

\usepackage{amsmath,amsfonts,bm}

\def\eqref#1{equation~\ref{#1}}

\def\1{\bm{1}}

\DeclareMathAlphabet{\mathsfit}{\encodingdefault}{\sfdefault}{m}{sl}
\SetMathAlphabet{\mathsfit}{bold}{\encodingdefault}{\sfdefault}{bx}{n}

\usepackage{hyperref}
\usepackage{url}
\usepackage{graphicx}
\usepackage{tcolorbox}
\tcbuselibrary{skins}

\usepackage{listings}
\definecolor{lightgray}{gray}{0.95}
\lstdefinestyle{myprompt}{
    basicstyle=\ttfamily\fontsize{7pt}{8pt}\selectfont,
    frame=none,
    breaklines=true,
    backgroundcolor=\color{lightgray},
    breakatwhitespace=true,
    breakindent=0pt,
    escapeinside={(*@}{@*)},
    numbers=none,
    numbersep=5pt,
    xleftmargin=5pt,
}
\tcbset{
  myaibox/.style={
    top=10pt,
    colback=white,
    colframe=black,
    colbacktitle=black,
    enhanced,
    center,
    attach boxed title to top left={yshift=-0.1in,xshift=0.15in},
    boxed title style={boxrule=0pt,colframe=white,},
  }
}
\newtcolorbox{AIbox}[2][]{myaibox,title=#2,#1}

\title{HuixiangDou: Overcoming Group Chat Scenarios with LLM-based Technical Assistance}

\author{%
  \textbf{Huanjun Kong}\qquad \textbf{Songyang Zhang} \qquad \textbf{Jiaying Li} \qquad \textbf{Min Xiao} \qquad \textbf{Jun Xu} \qquad \textbf{Kai Chen} \\ 
  \quad \\
  Shanghai AI Laboratory \\
}

\begin{document}

\maketitle

\begin{abstract}
In this work, we present HuixiangDou\footnote{HuixiangDou is a dish from the Chinese classical story 'Kong Yiji'. It's a meal Kong Yiji would routinely order from a local tavern, reflecting his humble conditions yet exalted spirit.}, a technical assistant powered by Large Language Models (LLM). This system is designed to assist algorithm developers by providing insightful responses to questions related to open-source algorithm projects, such as computer vision and deep learning projects from OpenMMLab. We further explore the integration of this assistant into the group chats of instant messaging (IM) tools such as WeChat and Lark. Through several iterative improvements and trials, we have developed a sophisticated technical chat assistant capable of effectively answering users' technical questions without causing message flooding. This paper's contributions include: 1) Designing an algorithm pipeline specifically for group chat scenarios; 2) Verifying the reliable performance of text2vec in task rejection; 3) Identifying three critical requirements for LLMs in technical-assistant-like products, namely scoring ability, In-Context Learning (ICL), and Long Context. We have made the source code, android app and web service available at \href{https://github.com/internlm/huixiangdou}{Github}, \href{https://openxlab.org.cn/apps/detail/tpoisonooo/huixiangdou-web}{OpenXLab} and \href{https://youtu.be/ylXrT-Tei-Y}{YouTube} to aid in future research and application. HuixiangDou is applicable to any group chat within IM tools.
\end{abstract}

\section{Introduction}

Authors of open-source projects often set up user groups on IM tools(like WeChat, Slack, Discord, etc.) for discussing project-related technical questions. As the number of users gradually increases, the maintainers, aiming to reduce the time spent on answering user questions while ensuring these questions are addressed, tend to pin some content or set up a bot to automatically answer FAQs. However, user inquiries are strongly correlated with their local development environments, and most messages in the group are unrelated to the project. However, traditional NLP solutions can neither parse the users' intent nor often provide the answers they desire.

ChatGPT, a large language model service from OpenAI, has a good performance in multiple test sets and natural language communication. However, directly integrating ChatGPT into group chats could lead to more severe issues.
\begin{itemize}
    \item ChatGPT is designed for single-user chat. If it responds too many messages within a group, it may impact others' experience and cause them to leave the group.
    \item For really valuable queries such as code implementation principles and modification methods, ChatGPT fails to provide correct answers. This is because its training data comes from the public internet, not domain-specific knowledge, and its data cannot be updated immediately with code modifications.
    \item Even though ChatGPT exhibits high accuracy in numerous datasets, it still faces the issue of hallucination. For example, asking "Who is the author of ncnn?" can yield an incorrect response related to "Nvidia Compute Library".
\end{itemize}

Hence, a technical assistant operating in group chats has different requirements.

\paragraph{Target true help-seekers} Technical assistant should not respond to non-technical content, such as politics, chit-chat, or personal information. They are only activated to provide responses to technical inquiries when users genuinely require assistance. 
\paragraph{Strictly no hallucination} Even a single instance of hallucination could make users perceive the bot as unreliable from a product perspective. Therefore, the system is implemented to avoid creating any false impressions of understanding.
\paragraph{Understand domain-specific knowledge} Possessing exclusivity not found in the public internet is the fundamental value of an assistant. Simultaneously, it can update the version content of the knowledge base at a relatively low cost.
\paragraph{No rush for response} Users might ask questions late at night without having much expectation for response time. Therefore, we can adopt more complex processing procedures.

\section{Approach}

\begin{figure}[ht]
\centering
\includegraphics[width=0.3\textwidth]{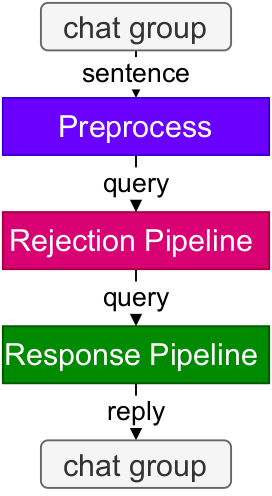}
\caption{\label{fig:approach}The overall structure of the approach. After the user's message is preprocessed, small talk will be filtered out, and only genuine questions will be responded to.}
\end{figure}

In addressing these unique needs, we started with a basic version and arrived at our current solution after making two improvements. As show in Figure \ref{fig:approach}, our final version consists of three parts: \textbf{Preprocess}, \textbf{Rejection} and \textbf{Response}.

The underlying philosophy of HuixiangDou is to eliminate irrelevant noise to improve precision; enhance retrieval capabilities to increase recall.

\subsection{Preprocess User Input}
In a chat group, multiple users may pose questions and communicate among themselves. However, the typical LLM chat template only support three roles: \textbf{system}, \textbf{user} and \textbf{bot}. 
Hence, we concatenate groupid and userid as the unique ID for users, in order to  
accommodate chat template.

Given that users are unlikely to describe their problem completely in one go, we pack multiple consecutive messages into a single one. In the process, we use OCR service to parse image and disregard other elements such as video, emojis, and voice messages.

Furthermore, extremely short messages that do not pose algorithmic challenges and messages that clearly do not seek interaction with the assistant by quoting others will also be disregarded.

\subsection{Rejection Pipeline}

\begin{figure}[ht]
\centering
\includegraphics[width=0.5\textwidth]{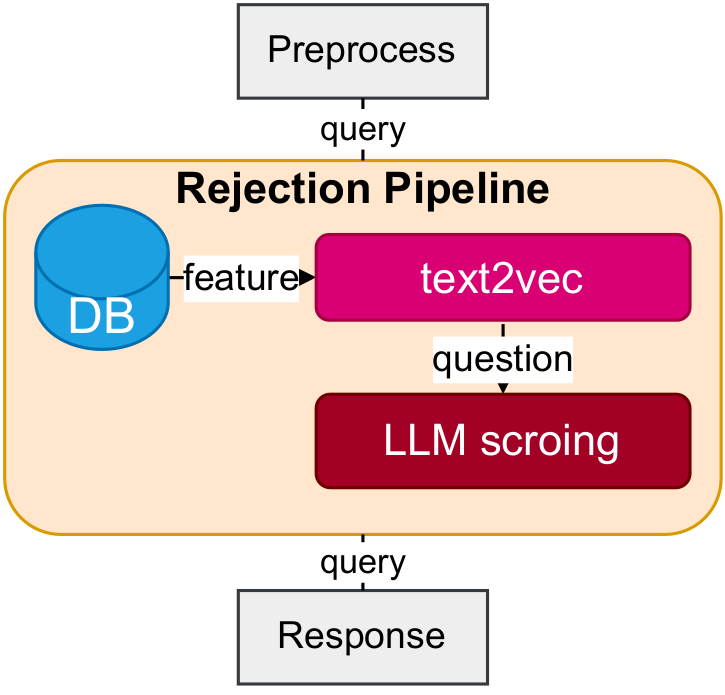}
\caption{\label{fig:rejection}The structure of rejection pipeline. We build a two-stage refusal-to-answer filter using text2vec and LLM scoring.}
\end{figure}

In group chat scenario, hallucinations come from two parts: user gossip and the model itself (where the model's training data and domain knowledge are not aligned). Rejections pipeline is designed for dismissing causal chat-like discourse, as shown in Figure \ref{fig:rejection}.

\paragraph{Refusal to Answer Based on Text2Vec}
LangChain \citep{LangChain} and wenda \citep{wenda} were originally used for RAG. After repeated tests, we think their retrieval abilities are normal, but surprisingly suitable for telling whether the question deserves to be answered. Undesired question are topics that are too distant from the knowledge base.

\paragraph{Refusal to Answer Based on LLM Scoring}
Because the text2vec model judges topic similarity, it's easy to be influenced by tone words in the chat group question. From the perspective of the text2vec model, there is a high degree of similarity between \textbf{ 'This development board is very good'} and \textbf{'This board is poorly designed'}. Influenced by moral and other factors, humans do not believe that these two sentences express the same meaning.

\subsection{Response Pipeline}

\begin{figure}
\centering
\includegraphics[width=0.5\textwidth]{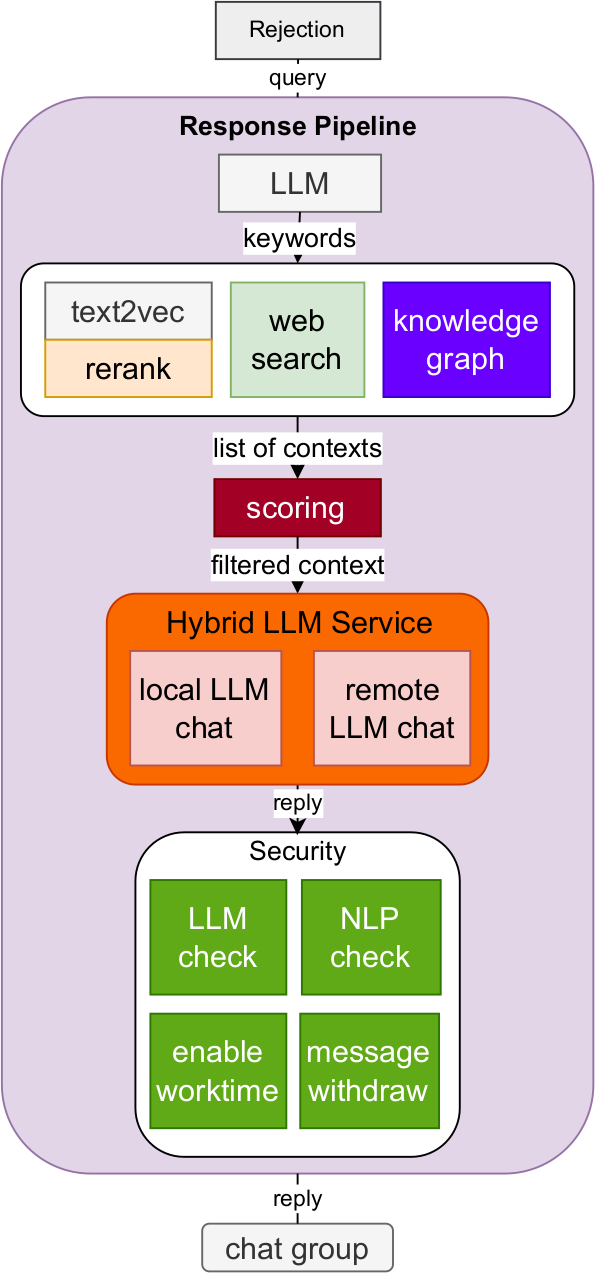}
\caption{\label{fig:response}The structure of response pipeline. We employ scoring and partial ordering to filter high-quality text from rerank model, web search and knowledge graph for the LLM to generate responses. To save costs, we mix and schedule different LLMs. We have established a set of security mechanisms to ensure that replies to chat groups do not involve sensitive topics.}
\end{figure}

Obviously, models with robust In-Context Learning capabilities can assuredly mitigate internal hallucinations through search mechanisms. As shown in Figure \ref{fig:response}, response Pipeline is designed to identify the underlying background knowledge of the issue.

The key here is to determine the importance of the information source and use them in order.

\paragraph{Extract Keywords}
User queries often contain many modal words, which can greatly impact the precision of text2vec models. Therefore, we cannot directly use original queries for text2vec search. As LLM excels at NLP part-of-speech segmentation tasks, we leverage it to extract keywords and phrases from the query.

\paragraph{Feature and Rerank}
We mixedly use LangChain and BCEmbedding \citep{bcembedding} to retrieve domain-specific knowledge. In this scenario, our search result is a list of document snippets. To fully utilize the context length of the model, we also employ LLM scoring to judge the relevance between the query and the document. This helps avoid any distractions for the LLM from irrelevant inputs. It's evident that LLM's In-Context Learning capability is extremely crucial for this scenario.

Due to the varying performance of different text2vec models, the Response Pipeline does not share a feature database with the Rejection Pipeline.

\paragraph{Web Search}
We first retrieve multiple search results (depends on the maximum token length supported by the model) and then utilize LLM scoring to filter the results associated with the question. These results are finally packed into the background document. This will return illegal content, so safety filtering is also necessary.

As stated by \cite{hsieh2024ruler}, although many models claim to perform excellently in needle-in-a-haystack task, their true long context ability remains valuable. Thus we can't feed all documents into the LLM at one time. This step involves deciding which pieces of information will make up the final input for the LLM Chat based on prior knowledge. For instance, for PyTorch-related questions, we tend to look up the official PyTorch documentation rather than some tech blog.

The trick here is that the data quality from web searches is not controllable and necessitates stringent review.

\paragraph{Knowledge Graph}
Search engines face the entire spectrum of internet information, but the background information implicit in the group chat technical assistant isn't fully utilized. For instance, users wouldn't actually ask mmdetection \citep{mmdetection} questions in the opencompass \citep{2023opencompass} user group. 

Based on sourcegraph \citep{mccoll2013brief}, we built a unique search engine for each repository, routing queries from different groups accordingly. This improvement enables the assistant to answer difficult questions that internet searches can't locate, which we discuss further in our LLM paging experiments. 

\paragraph{Scoring} The string of LLM's responses cannot be directly integrated into Python or Java, hence we implement process control through LLM scoring, such as for intent recognition and relevance assessment. The Experiments section will showcase more examples.

\paragraph{Hybrid LLM Service}
Our product focuses more on cost-efficiency and does not insist on a single model possessing all capabilities. We treat LLM chat as an RPC (Remote Procedure Call) service that can internally integrate multiple models for use as needed. 

The hybrid service is not a mishmash, it requires identifying the strengths of various models first and then invoking them based on the circumstances. In HuixiangDou, we can fully leverage Internlm2's scoring capability and kimi chat's long context ability.

\paragraph{LLM Response}

Directly using snippet to answer questions can lead to local optima. We read the original text corresponding to the snippet and hand it over to the LLM for processing along with the original question. The experimental part will showcase our work on Long Context.

Ultimately, we use LLM scoring to evaluate the relevance between the response and the query. If the relevance is low, assistant will not respond.

\paragraph{Security}
Many regions emphasize the security of AI applications. To ensure foolproof safety, we implemented four seat belts:

    \noindent $\bullet$ Check all string variables and their association with prohibited topics based on LLM scoring to prevent the generation of illegal content.
    
    \noindent $\bullet$ Integrate traditional security service to check whether the assistant's responses are illegal.
    
    \noindent $\bullet$ Set working hours for the assistant to ensure all activities are under human supervision.
    
    \noindent $\bullet$ Everyone can withdraw HuixiangDou's response if they deem inappropriate.

\section{Experiments}

In this section, we validate the feasibility of key technical points during the iterative process of the pipeline. Section \ref{subsec:fine-tuned-model} presents the fine-tuning process and conclusions of the LLM. Section \ref{subsec:rejection-pipeline} demonstrates Rejection Pipeline effects. Section \ref{subsec:llm-scoring} details the implementation and testing conclusions of the scoring method. Section \ref{subsec:long-context} is dedicated to the necessary experiments with Long Context responses. Section \ref{subsec:llm-paging} is an attempt to further enhance the search capabilities.

\subsection{Fine-tuned Model}
\label{subsec:fine-tuned-model}

\paragraph{Base model selection} Due to resource limitations, we cannot train from scratch and must select a base model for fine-tuning. Our selection criteria are as follows.

\noindent $\bullet$ Understanding domain-specific terminologies. That means training data includes the vocabulary needed for business operations. Otherwise, we believe that the required results cannot be calculated using attention score.
    
\noindent $\bullet$ Long context. Since we can use ReRoPE \citep{kexuefm9708} or dynamic NTK \citep{DynamicNTK} for expansion, a model supporting RoPE can be considered capable of handling long context.
    
\noindent $\bullet$ In-Context Learning (ICL) and stable scoring ability. 

\paragraph{Data preparation} Our training data comprises 28,000 QA pairs, which are made up of three parts:

1. During clean existing OpenMMLab group chat data, we removed personal information, and divided the dialogues into the QA format required for training. Among them, there are about 8,000 question-answer pairs.

2. For unanswered questions, we constructed responses using a larger LLM. These account for approximately 12,000 of the total.

3. We also scraped closed issues from GitHub, amounting to about 8,000 entries.

\paragraph{Train and test} We used the XTuner \citep{2023XTuner} qLoRA method to fine-tune on the 7B and 13B models respectively. Our learning rate is 2e-5 and epoch is 5. Regardless of the combination, there were significant issues with hallucinations. In the best version, the model learned colloquial expressions from users in WeChat groups, rather than technical answers. As shown in Appendix \ref{sec:finetuned-llm-examples}.

We believe that the biggest issue lies in data quality. Other users answer the questions casually and unprofessionally. In terms of domain-specific questions, the answers provided by LLM are not correct.

\subsection{Text2vec in Rejection Pipeline}
\label{subsec:rejection-pipeline}
We manually annotated hundreds of user contents, with human judgement determining whether they were related to domain-specific knowledge. We then used different text2vec models to construct a database and test the accuracy of refusal to answer. See table \ref{tab:my_label}.

\begin{table}
    \centering
    
\begin{tabular}{lccc}
\hline
\textbf{Model} & \textbf{Precision} & \textbf{Recall}\\\hline
text2vec-large-chinese & 0.99 & 0.92 \\
text2vec-bge-large-chinese & 0.95 & 0.81 \\\hline

\end{tabular}
\caption{Test the refusal to answer using different text2vec models on manually annotated data. text2vec model has demonstrated strong robustness in the refusal-to-answer task.}
\label{tab:my_label}
\end{table}

We also examined the impact of various text split methods on precision, including \texttt{langchain.MarkdownHeaderTextSplitter}, \texttt{langchain.CharacterTextSplitter} and their combined implementation. Experiments showed that the impact of the split method on the precision of refusal to answer can be disregarded.

\subsection{LLM Scoring in Intent Recognition}
\label{subsec:llm-scoring}
LLM Scoring has been utilized in intent determination, the evaluation of relevance between questions and background, as well as within security contexts.

This is implemented by determining the final score of the task. In engineering practice, we often use integers and booleans instead of strings to determine the result of an if statement. Appendix~\ref{sec:scoring-prompt} is the LLM scoring prompt.

\begin{figure}[ht]
\centering
\includegraphics[width=0.8\textwidth]{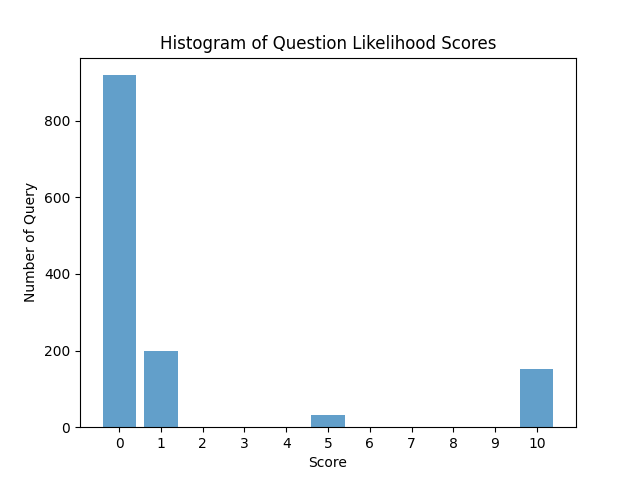}
\caption{\label{fig:scoring-distribution}Question likelihood with InternLM2-7B on 1,303 domain-related group chat sentences, 11.6\% are questions. This distribution aligns with common sense, and the scoring method can effectively handle intent recognition tasks.}
\end{figure}

We randomly selected 1,303 domain-related queries and use InternLM2-7B to estimate the likelihood of the query being a question—the higher the score, the more likely it is a question. As shown in Figure \ref{fig:scoring-distribution}, 11.6\% of the content are user questions, which is in line with common sense. 

While elaborating on the problem can enhance the model's performance, the model's perspective of the world doesn't straightforwardly equate to that of a human. For instance, a submarine can "swim," but this is not equivalent to human swimming.

We extracted 11,362 sentences from the content sent by users. To improve the precision of rejection pipeline, we included scoring examples in the prompt, see Appendix \ref{sec:scoring-prompt-with-example}. However, after adding these examples, the scores for all 7,753 pieces of data increased.

But in reality, more than 80\% of group chat consists of idle chatter. If precision improves, the score distribution should present a polarized state.

\subsection{Long Context Optimization}
\label{subsec:long-context}
Based on our experience in Chinese-English bilingual scenarios, the token length of a search result can exceed 11k. Considering the prompt and historical dialogue, the model's max token length should be more than 16k. Only with 32k can we achieve relatively good results.

Considering the prohibitive training cost of YaRN \citep{peng2023yarn}, we optimized ReRoPE's inference \footnote{See https://github.com/InternLM/lmdeploy/pull/625} performance using Triton \citep{2019triton}, also introducing dynamic quantization\footnote{See https://github.com/InternLM/lmdeploy/pull/718} base on LMDeploy \citep{2023LMDeploy}. 

\begin{table}[h]
    \centering
    
\begin{tabular}{lcc}
\hline
\textbf{Model} & \textbf{Length(Memory)} & \textbf{Precision}\\
\hline
\verb|baseline| & 4k(*65GB) & 1.0 \\
\verb|ReRoPE| & 14k(79GB) & 1.0 \\
\verb|ReRoPE+Quant| & 40k(75GB) & 1.0 \\\hline

\end{tabular}

\caption{Passkey test results of using different methods on openbuddy-llama2-13B-v8.1-fp16. To optimize speed, LMDeploy automatically pre-allocates memory based on the GPU, hence the base version occupy 65GB. We ultimately achieved 40k long text on a single card, proving that the ReRoPE method is feasible.}
\label{tab:passkey}
\end{table}

Eventually, we achieved support for 40k token length on an A100 80G card. Table \ref{tab:passkey} is our precision test report for passkey retrieval, with the base model being openbuddy-llama2-13B-v8.1-fp16.

\section{Other Attempts}

In order to enhance the accuracy, we have explored Natural Language Processing (NLP) as well as prompting techniques, but these methods have insurmountable shortcomings, and thus were ultimately not adopted.

\subsection{NLP in RAG}
Since the capabilities of the text2vec model are limited, we have tried to simplify the query and document with NLP methods. For example, inputting "How to install mmdet and mmcv" will identify CC part of speech, thereby decomposing into two simple questions. But in actual operation, we encountered more difficult problems.

\noindent $\bullet$ Domain-specific part-of-speech tagging lacks precision. For example, in the field of deep learning, the part of speech for "deploy" depends on the context, which is different from daily communication.

\noindent $\bullet$ Bilingual problems. HanLP \cite{he-choi-2021-stem} exhibits subpar performance in English, and other well-known projects do not support Chinese. Utilizing translation APIs to bridge this gap in bilingual models poses further complications. Due to the lack of appropriate translations for certain terms, it can result in significant misinterpretations, such as with the term "transformers".

\subsection{Prompt Technics}

\label{subsec:llm-paging}
\paragraph{Paging} Suppose we want to make LLM understand an entire repository via prompts. Even the latest 192k context length can't accommodate the full source code of OpenCompass. During ReRoPE optimization, we also realized that the transformer kv cache and attention score mechanism severely limit the maximum context length.

Inspired by the operating system paging mechanism, we compressed the Python module into a single description, thereby shrinking the OpenCompass project within 120k. For user technical queries, we let LLM decide which modules to view, then extract the module source code for secondary inquiries. However, in practice, LLM only finds partial source code using a 128k context, and user questions may involve multiple knowledge points. Appendix \ref{sec:llm-paging-example} is an LLM Paging example without any web search nor RAG results.

\paragraph{Rephrase and Respond} \cite{deng2023rephrase} attempts to enhance the prompt using LLM, but this is constrained by the understanding ability of the base model, making it incapable of extending this technique to interrogative sentences. Otherwise, it would lead to confusion in the LLM. Here is an scoring example.

\begin{figure*}[ht] 
\vspace{-5mm}
\begin{AIbox}{Rephrase and Respond Example}
{\color{orange}\bf User:} \\
{
"Determine whether the following sentences are topical interrogative sentences, with results ranging from 0 to 10.
Provide scores directly without explanation."\\
Rephrase and expand the question, and respond.
}\\ 
{\color{blue}\bf Assistant:} \\
{
1
}
\end{AIbox} 
\caption{RaR prompt is not applicable to interrogative sentences.}
\label{fig: rar example}
\end{figure*}
\paragraph{ReAct} \cite{yao2023react} utilizes training data to potentially generate fixed-format json results based on inputs, which are then employed to invoke tools such as search engines. However, from a product perspective, being unable to debug indicates a significant risk. In practical use, search behaviors are often triggered even for simple queries. Considering that this approach requires training data, we don't deem it cost-effective for practical use.

\section{Conclusion and Limits}
In this work, we demonstrated the feasibility of using text2vec for refusal response, and multiple search methods can substantially mitigate the hallucination caused by LLMs.

As long as an LLM has the following capabilities, instead of strange prayer postures, it can sufficiently address most demands within group chat scenarios:

    $\bullet$ Understanding domain-specific terminologies.
    
    $\bullet$ Supporting a minimum token length of 16k.
    
    $\bullet$ Scoring capability.
    
    $\bullet$ In-Context Learning.

However, as users' questions become more professional, it's increasingly difficult to provide satisfactory responses based on the prompt and search method. This necessitates that the LLM truly understands the source code in the repository. We think efficient further pretrain is the next stage solution.

Due to the limitations of the ChatML \citep{ChatML} format, we have merely divided the group messages according to the user, which in fact has led to a significant loss of contextual information. The new chat format should fully expressing the context of the problem, the historical messages of the speaker, and the remarks.

Additionally, users are very fond of first sending log screenshots before asking questions. Many valuable contexts are contained within these images, but HuixiangDou does not support multimodal. We will work on it.

\section{Acknowledgments}
    \noindent $\bullet$ We would like to express our gratitude towards the OpenMMLab users and \href{https://github.com/Tencent/ncnn/graphs/contributors}{ncnn contributors} for their understanding and tolerance of the numerous bugs in the technical assistant. 
    
    \noindent $\bullet$ We are grateful to the teams at OpenCompass, XTuner, and LMDeploy for their guidance during the exploratory phase of the project.
    
    \noindent $\bullet$ Our thanks also go to \href{https://www.moonshot.cn/}{Moonshot AI} and \href{https://github.com/xxr3376}{Xinran Xu} for providing a free 128k context LLM API.
    
    \noindent $\bullet$ We extend our appreciation to \href{https://spaces.ac.cn/}{Jianlin Su}, the author of RoPE, for his profound insights into the structure of transformers.
    
    \noindent $\bullet$ Finally, we want to thank \href{libowen@pjlab.org.cn}{Bowen Li} and \href{liukuikun@pjlab.org.cn}{Kuikun Liu} for their ideas on NLP, thank \href{jasonyso@aliyun.com}{Song Yang} for his android app contribution on Github, thank \href{huwenxing@pjlab.org.cn}{Wenxing Hu} for his method of integrating WeChat and \href{mailto:2806651841@qq.com}{Siyue Zhao} for the proofreading on this Report.

\bibliography{iclr2024_conference}
\bibliographystyle{plainnat}

\newpage
\appendix

\section{Fine-tuned LLM Chat Examples}
\label{sec:finetuned-llm-examples}
\begin{figure*}[ht!]
\vspace{-5mm}
\begin{AIbox}{Fine-tuned LLM Chat Examples}
{\color{orange}\bf User:} \\
{
Can mmdet3.0 use version 0.7 of mmcv ?
}\\ 
{\color{blue}\bf Assistant:} \\
{
mmdetection3.0 and mmcv version 0.7 are not compatible due to differences in certain features. Mmdetection 3.x requires mmcv 2.0 or a higher version. If you need to use an older version of mmcv, you may consider downgrading to mmdetection 2.x.
}\\
\\
{\color{red}\bf Hallucination Example:}\\
{\color{orange}\bf User:}
{
Does mmdet3.0 support mmcv0.7 ?
}\\ 
{\color{blue}\bf Assistant:}
{
There's no need for that, the version of mmcv should be between 1.3.3 and 1.5.0.
}\\
\\
{\color{red}\bf Response Style Example:}\\
{\color{orange}\bf User:} 
{
mmyolo uses the coco format, right ?
}\\ 
{\color{blue}\bf Assistant:} 
{
Haha, your understanding is very precise. Yes, mmyolo indeed utilizes the coco format.
}
\end{AIbox} 
\caption{Fine-tuned LLM examples.}
\label{fig:finetune}
\end{figure*}

\section{System prompt in Scoring}
\label{sec:scoring-prompt}
\begin{figure*}
\begin{AIbox}{System prompt in Scoring}
{\bf Prompt:} \\
{
Determine whether the following sentences are topical interrogative sentences, with results ranging from 0 to 10.
Provide scores directly without explanation.

\begin{lstlisting}[style=myprompt]
Scoring standards:
A score of 10 for sentences with subject, predicate, and object that are interrogative;
points deducted for missing subject, verb, or object; a score of 0 for declarative sentences;
a score of 0 for non-interrogative sentences.
\end{lstlisting}

New question "\{\}", what is the score? Provide scores directly without explanation.
}
\end{AIbox} 
\caption{Prompt for LLM scoring task.}
\end{figure*}

\section{System Prompt with Examples}
\label{sec:scoring-prompt-with-example}
\begin{figure*}
\begin{AIbox}{LLM Scoring with Examples}
{\bf Prompt:} \\
{
Determine whether the following sentences are topical interrogative sentences, with results ranging from 0 to 10.
Provide scores directly without explanation.

\begin{lstlisting}[style=myprompt]
Scoring standards:
A score of 10 for sentences with subject, predicate, and object that are interrogative;
points deducted for missing subject, verb, or object; a score of 0 for declarative sentences;
a score of 0 for non-interrogative sentences.
\end{lstlisting}

\begin{lstlisting}[style=myprompt]
Here are some examples:
Question "Excuse me, how should mmdeploy be installed?", Score: 9
Question "How to apply for modification of rpm?", Missing subject, Score: 7
Question "Please check if the environment is installed with your revised version", It's a declarative sentence, Score: 0
Question "If you treat bot as a living person, everything will go smoothly", It's not a question, Score: 0
\end{lstlisting}

New question "\{\}", what is the score? Provide scores directly without explanation.
}
\end{AIbox} 
\caption{LLM scoring with examples, build prompt with examples does not enhance the score.}
\end{figure*}

\section{LLM Paging Example}
\label{sec:llm-paging-example}
\begin{figure*}
\vspace{-5mm}
\begin{AIbox}{LLM Paging Example}
{\color{orange}\bf User:} \\
{
How do I output the results of triviaqa 5shot in a summarizer?
}\\ 
{\color{olive}\bf Real Solution:} \\
{
 Add an element: \texttt{'triviaqa\_5shot'} to the array \texttt{summarizer['dataset\_abbrs']} in the configuration file.
}\\
{\color{blue}\bf Assistant:} \\
{
First, you need to ensure that the results of "triviaqa 5shot" are stored in your code in an appropriate data structure (such as a dictionary or list), and that these results can be accessed from your data processing or loading section.

Next, you need to find or create a method within the relevant summarizer class that handles and formats the "triviaqa 5shot" results. This might involve parsing, computing, or transforming data so they can be correctly included in the output of the summarizer.

Finally, you need to insert code into the output methods of the summarizer (like \_form\_table or \_format\_raw\_txt) so that the "triviaqa 5shot" results are included when generating summaries. This might involve adding new columns or rows to a table or text, or modifying existing data to include new information.
}
\end{AIbox} 
\caption{LLM Paging only searched the source code of \texttt{opencompass.summarizer}, and was not aware of the existence of the config file.\\}
\end{figure*}

\end{document}